\documentclass[conference]{IEEEtran} 

\usepackage{balance}
\usepackage{graphicx}
\usepackage{amsmath}
\usepackage{amssymb}
\usepackage{booktabs}
\usepackage{float}
\usepackage{stfloats}
\usepackage{url}
\usepackage{xurl}
\usepackage{hyperref}
\hypersetup{
    colorlinks=true,
    linkcolor=black,
    urlcolor=black,
    citecolor=black
}
\usepackage{algorithm}
\usepackage{algorithmic}
\usepackage{caption}
\usepackage{placeins}
\usepackage{needspace}

\title{Agent-Based Modular Learning for Multimodal Emotion Recognition in Human-Agent Systems}

\author{
    \IEEEauthorblockN{
        Matvey Nepomnyaschiy$^{1,2}$ \qquad
        Oleg Pereziabov$^{1}$ \qquad
        Anvar Tliamov$^{1}$
    }
    \\[-0.5em]

    \IEEEauthorblockN{
        Stanislav Mikhailov$^{2}$ \qquad
        Ilya Afanasyev$^{1,3}$
    }
    \\[0.5em]

    \IEEEauthorblockA{
        $^1$ Research Center of the Artificial Intelligence Institute, Innopolis University, Innopolis, 420500, Russia \\
        $^2$ ITMO University, St. Petersburg, 197101, Russia \\
        $^3$ Saint Petersburg Electrotechnical University ``LETI'', St. Petersburg, 197022, Russia
        \\[1.0em]
        Emails: \\
        m.nepomnyashchy@innopolis.ru,\; m.nepomnyashchy@niuitmo.ru\; (Matvey) \\
        o.perezyabov@innopolis.ru\; (Oleg) \\
        a.tliamov@innopolis.university\; (Anvar) \\
        stasmikhailov@niuitmo.ru\; (Stanislav) \\
        i.afanasyev@innopolis.ru,\; imafanasev@etu.ru\; (Ilya)
    }
}

\newcommand{\BibTeX}{\rm B\kern-.05em{\sc i\kern-.025em b}\kern-.08em\TeX}

\begin{document}

\maketitle

\begin{abstract}
Effective human-agent interaction (HAI) relies on accurate and adaptive perception of human emotional states. While multimodal deep learning models—leveraging facial expressions, speech, and textual cues—offer high accuracy in emotion recognition, their training and maintenance are often computationally intensive and inflexible to modality changes. In this work, we propose a novel multi-agent framework for training multimodal emotion recognition systems, where each modality encoder and the fusion classifier operate as autonomous agents coordinated by a central supervisor. This architecture enables modular integration of new modalities (e.g., audio features via emotion2vec), seamless replacement of outdated components, and reduced computational overhead during training. We demonstrate the feasibility of our approach through a proof-of-concept implementation supporting vision, audio, and text modalities, with the classifier serving as a shared decision-making agent. Our framework not only improves training efficiency but also contributes to the design of more flexible, scalable, and maintainable perception modules for embodied and virtual agents in HAI scenarios.
\end{abstract}

\IEEEkeywords Multi-Agent Systems, Emotion Recognition, Multimodal Learning, Modular Architecture, Supervisor Architecture, Agent Coordination, Human-Agent System
\endIEEEkeywords

\section{Introduction}

Human-agent interaction (HAI) is becoming increasingly important as autonomous agents need to understand and respond to human emotions to work effectively with people~\cite{Breazeal2003}. To achieve this, agents must be able to recognize emotions from multiple sources such as facial expressions, speech, and text~\cite{Picard1997}. This multimodal emotion recognition capability is essential for creating socially intelligent agents that can adapt their behavior based on human emotional states.

Current approaches to multimodal emotion recognition typically use large neural networks that process all input types together~\cite{Baltruvsaitis2018}. While these methods work well in controlled environments, they face several problems in real-world applications. These monolithic systems are computationally expensive to train, difficult to modify when new input types are added, and challenging to maintain because changing one component affects the entire system~\cite{Ramachandran2019}. Multi-agent systems offer a promising alternative approach to solving complex problems by dividing tasks among specialized agents~\cite{Wooldridge2009}. For emotion recognition, this means that each input type (like facial expressions or speech) can be processed by a dedicated agent with specific expertise~\cite{Stone2000}. These agents can then coordinate their results, making the system more modular and easier to update or improve individual components.

Recent developments in machine learning have produced powerful models for emotion recognition, such as emotion2vec~\cite{Ma2024}, which works well across different languages and audio conditions. Similarly, specialized models for face detection~\cite{Xu2024}\footnote{\url{https://github.com/YapaLab/yolo-face}} and text emotion analysis\footnote{\url{https://huggingface.co/ai-forever/FRIDA}} have achieved excellent performance in their specific areas. However, combining these different models into a single emotion recognition system remains challenging, especially considering the computational costs and design limitations of traditional approaches.

In this work, we propose a multi-agent framework for multimodal emotion recognition that addresses these limitations using a supervisor-based architecture. Our approach processes video input through specialized agents for each primary modality (facial expressions, speech, and text), with an additional Audio Event Detection (AED) component that provides auxiliary audio tags (e.g., speech presence) rather than a separate modality in the fusion space. A central supervisor then coordinates these agents and makes the final emotion prediction. This design allows for easy integration of new input types, simple replacement of outdated components, and reduced computational costs during training and operation. The complete implementation is available as open-source software\footnote{\href{https://github.com/MatNepo/MMER_MultiAgent}{https://github.com/MatNepo/MMER\_MultiAgent}}.

The main contributions of this work are: (1) we introduce a multi-agent architecture for multimodal emotion recognition that processes video input through specialized agents using supervisor-based coordination to overcome the limitations of traditional approaches; (2) we demonstrate our approach through a complete implementation that supports vision, audio, and text processing extracted from video, using state-of-the-art models including emotion2vec, YOLOv8-Face, and FRIDA; and (3) we provide a modular framework that improves system flexibility and maintainability for real-world applications, allowing for easy updates and improvements without affecting the entire system. Our multi-agent framework provides a new approach to multimodal emotion recognition that offers better modularity and flexibility compared to traditional methods. While our current implementation uses logistic regression as the initial classifier, the modular design makes it easy to integrate more advanced classification methods in the future. This work contributes to the development of more robust and maintainable perception systems for human-agent interaction applications.

\section{Related Work}

Multimodal emotion recognition (MER) integrates signals from visual, acoustic, and textual modalities to improve affective understanding in human–agent interaction. Early research focused on unimodal systems such as ResNet and ViT for visual cues \cite{he2016deep,dosovitskiy2020image}, PANNs and Whisper Large V3 Turbo\footnote{\url{https://huggingface.co/openai/whisper-large-v3-turbo}} for speech \cite{kong2020panns,radford2023robust}, and RoBERTa for textual emotion modeling \cite{liu2019roberta}. While effective for isolated channels, these systems fail to capture inter-modal dependencies critical to robust emotion inference.

To address this, multimodal fusion models employ tensor fusion networks, gating mechanisms, or transformer-based fusion (e.g., BLIP-2) to integrate modalities \cite{zadeh2018multimodal,liu2022mmim}. However, most of these architectures are monolithic, with tightly coupled input streams, which limits their adaptability and interpretability. Updating or replacing a modality encoder often requires retraining the entire model \cite{han2021survey}.

Recent work explores modular and collaborative architectures inspired by multi-agent systems \cite{wooldridge2009introduction,weiss2013multiagent}. Studies such as Inside Out \cite{savchenko2024inside} and Project Riley \cite{ortigoso2025project} employ autonomous agents coordinated by supervisory modules, demonstrating enhanced modularity and interpretability in complex reasoning tasks. Yet, these systems still face unresolved challenges regarding coordination overhead, synchronization, and consistent performance across modalities \cite{yaseen2024yolov9,radford2023robust}.

In the domain of emotion recognition, a multi-agent architecture for multimodal emotion recognition was proposed \cite{raynova2020multiagent}, using a loosely-coupled framework with separate agents for visual, acoustic, and textual modalities coordinated by a central AgencyProcessor. This work employed classical techniques and BDI (Belief--Desire--Intention) agent concepts, but remained primarily conceptual without addressing practical implementation challenges. Our work extends this approach by implementing a practical multi-agent framework that leverages modern SOTA models (YOLOv8-Face, emotion2vec, Whisper, FRIDA) as specialized agents, replacing the abstract BDI framework with concrete neural network-based agents and a supervised fusion classifier orchestrator.

Prior works in emotion recognition mainly focus on fusion strategies rather than distributed collaboration or adaptive modality management \cite{hu2025beyond}. To our knowledge, no study has systematically evaluated how modular agent-based architectures affect predictive accuracy, retraining efficiency, and interpretability in MER.

Our work builds on these developments by proposing a distributed multi-agent framework where each modality-specific encoder operates as an autonomous agent. This design aims to preserve predictive performance while improving adaptability and transparency during modality evolution, addressing key limitations of prior monolithic MER architectures.

\section{Problem Formulation}

Let $\mathcal{V}$ be the input video data, from which we extract multiple modalities. In our framework, there are \emph{three primary modalities} used by the fusion classifier: facial expressions ($m_{fed}$) extracted from video frames, speech audio ($m_{ser}$) extracted from the video audio track, and transcribed text ($m_{ted}$) obtained from speech-to-text processing. In addition, we compute audio event tags ($z_{aed}$) from audio analysis using an Audio Event Detection (AED) component, but \textbf{AED is not treated as a separate modality in the fusion space} and is instead used as auxiliary metadata (e.g., speech presence) for gating and reliability estimation. The multimodal input used for fusion can thus be represented as:
\begin{equation}
\mathcal{M} = \{X_1, X_2, X_3\} = \{m_{fed}, m_{ser}, m_{ted}\}
\end{equation}
where $\mathcal{M}$ represents the multimodal input space, and each $X_n$ (for $n = 1, 2, 3$) corresponds to a primary modality. The AED-derived tags $z_{aed}$ act as side information that conditions the processing of $\mathcal{M}$ (e.g., enabling or disabling TED when speech is absent). The emotion recognition task can be formulated as learning a mapping function $f: (\mathcal{M}, z_{aed}) \rightarrow \mathcal{Y}$, where $\mathcal{Y}$ is the set of emotion classes $\{\text{neutral, joy, sadness,}$ $\text{surprise, fear, disgust, anger}\}$.

An effective emotion recognition system for HAI applications must satisfy several key requirements. First, individual modality processors should be independent and replaceable without affecting other components, ensuring modularity. Second, the system should easily accommodate new modalities or updated models, providing scalability. Third, processing should be computationally efficient for real-time applications. Fourth, the system should be able to process video input and automatically extract multiple modalities, making it suitable for real-world applications. Finally, the system should maintain performance across varying environmental conditions, ensuring robustness. These requirements motivate our multi-agent approach, where each modality is handled by a specialized agent that can be developed, updated, and maintained independently.

\section{Proposed Multi-Agent Architecture}

\begin{figure*}[t]
  \centering
  \includegraphics[width=\textwidth]{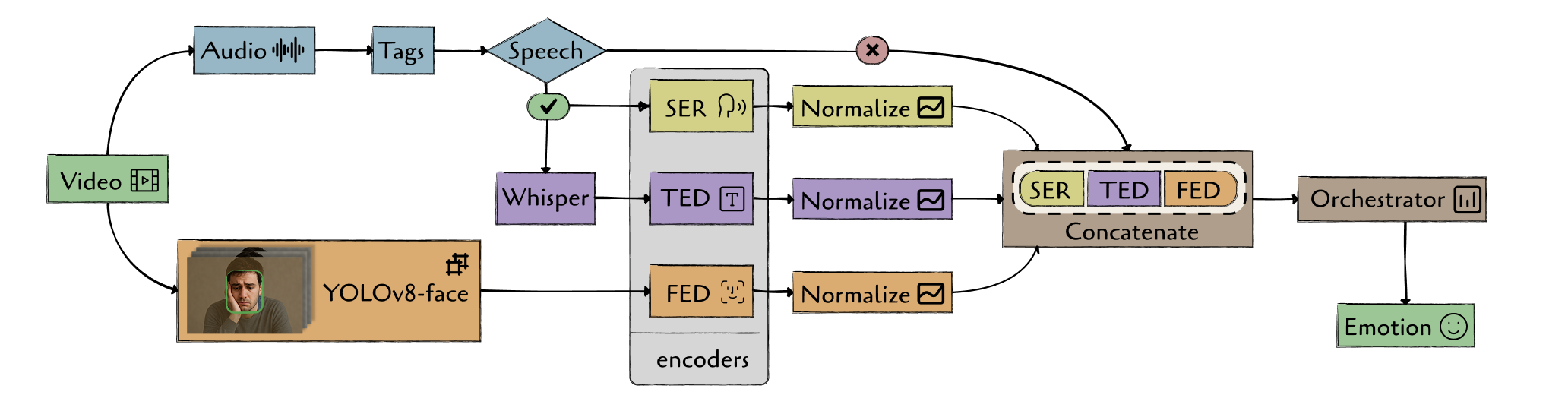}
  \caption{Multi-agent architecture for multimodal emotion recognition. 
  Three primary modalities are processed by specialized agents (FED: Facial Emotion Detection, 
  SER: Speech Emotion Recognition, TED: Text Emotion Detection), while AED (Audio Event Detection) 
  provides auxiliary audio tags such as speech presence and is \emph{not} treated as a separate 
  modality in the fusion space. A central supervisor coordinates the agents and makes the final decision.}
  \label{fig:architecture}
\end{figure*}

Our multi-agent framework processes video input through four specialized modality agents and a central supervisor classifier. The system first extracts audio and video frames from the input video, then processes each modality independently. Each modality agent is responsible for processing its specific input type and extracting relevant emotion features or predictions. The supervisor classifier coordinates these agents and makes the final emotion prediction. Figure~\ref{fig:architecture} illustrates the overall system architecture, showing the flow of information from video input through modality extraction, individual agents, to the final emotion classification.

While Figure~\ref{fig:architecture} presents the overall system architecture, Figure~\ref{fig:multimodal_agents} focuses specifically on the separation between modality agents and their supporting tools. The three modality agents—Audio, Text, and Image—are grouped under the Modality Agents block as autonomous components responsible for modality-specific reasoning and embedding generation. Complementary domain Tools provide deterministic preprocessing that supplies agents with clean, structured inputs (e.g., speech denoising and detection for Audio, Whisper-based transcription for Text, face extraction for Image). The orchestrator mediates the exchange of embeddings and coordinates fusion, but the emphasis of this figure is to clarify what the system considers agents versus tools.

\begin{figure}[t]
  \centering
  \includegraphics[width=0.9\columnwidth]{images/multimodal_agents.png}
  \caption{Operational pipeline highlighting the orchestrator, modality agents (Audio, Text, Image), and supporting tools.}
  \label{fig:multimodal_agents}
\end{figure}

\subsection{Facial Emotion Detection}

The Facial Emotion Detection Agent (FED) processes video frames extracted from the input video to detect and analyze facial expressions using YOLOv8-Face for face detection and ResNet-50~\cite{He2016} for emotion classification. The agent operates through a multi-stage pipeline that includes frame extraction, face detection using bounding boxes, and emotion classification for each detected face.

\textbf{Input:} Video frames extracted from the input video. \textbf{Output:} Variable-length sequences of 512-dimensional facial emotion embeddings for each detected face, with emotion class probabilities. The pipeline consists of three stages: (1) YOLOv8-Face detects faces in video frames and outputs bounding box coordinates; (2) ResNet-50 extracts facial features from cropped face regions (defined by bounding boxes); (3) emotion classification layer produces emotion probabilities from ResNet-50 features.

The key contribution of our FED agent is the direct emotion classification from facial features using ResNet-50. For each detected face, the emotion classification can be formulated as:
\begin{equation}
P_{fed}(c) = \text{softmax}(\mathbf{W}_{fed} \mathbf{f}_{face} + \mathbf{b}_{fed})
\end{equation}
where $\mathbf{f}_{face}$ represents the facial features extracted by ResNet-50, $\mathbf{W}_{fed}$ and $\mathbf{b}_{fed}$ are learnable parameters, and $c$ represents the emotion class.

\subsection{Speech Emotion Recognition}

The Speech Emotion Recognition Agent (SER) processes audio extracted from the input video to recognize emotions in speech using emotion2vec~\cite{Ma2024}. \textbf{Input:} Audio track extracted from the input video (WAV format). \textbf{Output:} Variable-length sequences of 256-dimensional audio emotion embeddings. The agent operates on the audio track extracted from the video and produces emotion predictions directly from the audio embeddings using the pre-trained emotion2vec model, which encodes speech into emotion-aware representations.

Our contribution lies in the direct emotion classification from audio embeddings using emotion2vec. The emotion classification is performed using a multi-layer perceptron with dropout regularization:
\begin{equation}
\begin{split}
P_{ser}(c) &= \text{softmax}(\mathbf{W}_{ser,2} \cdot \text{Dropout}(\text{ReLU}(\mathbf{W}_{ser,1} \mathbf{e}_{emotion} +\\
&\quad + \mathbf{b}_{ser,1}), p=0.3) + \mathbf{b}_{ser,2})
\end{split}
\end{equation}
where $\mathbf{e}_{emotion}$ represents the audio embeddings from emotion2vec, $\mathbf{W}_{ser,1}, \mathbf{W}_{ser,2}$ and $\mathbf{b}_{ser,1}, \mathbf{b}_{ser,2}$ are learnable parameters, and $p=0.3$ is the dropout probability.

\subsection{Text Emotion Detection}

The Text Emotion Detection Agent (TED) handles text-based emotion recognition through a two-stage process: speech-to-text transcription using OpenAI Whisper Large V3 Turbo~\cite{Radford2023} on the audio extracted from video, followed by text emotion analysis using FRIDA for emotion classification. \textbf{Input:} Audio track extracted from the input video. \textbf{Output:} Variable-length sequences of 768-dimensional text emotion embeddings. The pipeline consists of two stages: (1) Whisper transcribes speech audio into text tokens; (2) FRIDA processes the transcribed text and produces emotion-aware embeddings.

Our contribution focuses on the emotion classification approach for transcribed text. The emotion classification follows a similar pattern to the SER agent:
\begin{equation}
\begin{split}
P_{ted}(c) &= \text{softmax}(\mathbf{W}_{text,2} \cdot \text{ReLU}(\mathbf{W}_{text,1} \mathbf{e}_{text} + \\
&\quad + \mathbf{b}_{text,1}) + \mathbf{b}_{text,2})
\end{split}
\end{equation}
where $\mathbf{e}_{text} \in \mathbb{R}^{768}$ represents the text emotion embeddings from FRIDA, $\mathbf{W}_{text,1}, \mathbf{W}_{text,2}$ and $\mathbf{b}_{text,1}, \mathbf{b}_{text,2}$ are learnable parameters, $c$ represents the emotion class, and the text is obtained from Whisper Large V3 Turbo transcription of the video's audio track.

\subsection{Audio Event Detection}

The Audio Event Detection Agent (AED) analyzes audio events from the video's audio track to provide additional context for emotion recognition using CNN-14~\cite{Kong2017} trained on AudioSet~\cite{gemmeke2017audio}. \textbf{Input:} Audio track extracted from the input video (converted to spectrograms). \textbf{Output:} 527-dimensional audio event feature vector representing probabilities for 527 different audio event categories from AudioSet, including speech-related events (e.g., ``Speech'', ``Male speech, man speaking'', ``Female speech, woman speaking'', ``Conversation'', ``Shout'', ``Screaming'', ``Whispering'') and non-speech events (e.g., ``Music'', ``Animal sounds'', ``Vehicle sounds'', ``Nature sounds''). 
The agent produces audio event tags that are mapped to emotions, and in fusion classifier mode, it determines whether speech is present in the audio:

\begin{equation}
P_{aed}(c) = \text{softmax}(\mathbf{W}_{aed} \cdot \text{speech\_filter}(\mathbf{f}_{audio}) + \mathbf{b}_{aed})
\end{equation}
where $\mathbf{f}_{audio} \in \mathbb{R}^{527}$ represents the audio event features from CNN-14, $\mathbf{W}_{aed}$ and $\mathbf{b}_{aed}$ are learnable parameters, $c$ represents the emotion class, and the speech filter determines whether speech is present in the audio for fusion classifier mode by checking if any of the top-5 predicted audio event tags belong to speech-related categories.

\subsection{Feature Aggregation and Adapter Transformation}

A critical aspect of our multi-agent architecture is the feature aggregation and adapter transformation that enables effective fusion of embeddings from different modalities. Each modality agent produces embeddings of different dimensions and temporal lengths: FED produces variable-length sequences of 512-dimensional embeddings, SER produces variable-length sequences of 256-dimensional embeddings, and TED produces variable-length sequences of 768-dimensional embeddings. 

\begin{figure*}[t]
  \centering
  \includegraphics[width=\textwidth]{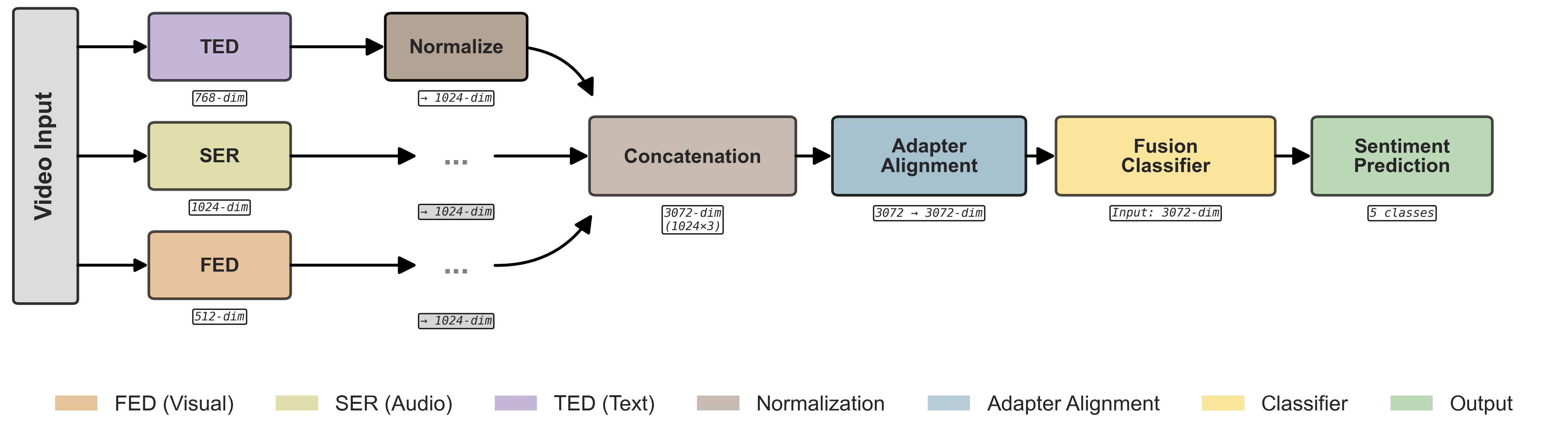}
  \caption{Dimension transformation pipeline showing the flow from video input through modality agents (FED, SER, TED) with different output dimensions, normalization to uniform 1024-dimensions per modality, concatenation, adapter alignment, and classification.}
  \label{fig:dimension_pipeline}
\end{figure*}

To enable effective fusion, we first aggregate temporal sequences from each modality into fixed-length vectors. For a modality $i$ with temporal sequence $\{\mathbf{f}_{i,t}\}_{t=1}^{T_i}$ of length $T_i$, we apply temporal pooling (mean, median, or max) to obtain a single embedding vector:

\begin{equation}
\mathbf{f}_{agg,i} = \text{pool}(\{\mathbf{f}_{i,t}\}_{t=1}^{T_i})
\end{equation}

where $\text{pool}$ denotes the temporal pooling operation, and $\mathbf{f}_{agg,i}$ is the aggregated embedding vector for modality $i$. We then normalize all aggregated embeddings to a uniform dimension of 1024 per modality using padding or truncation:

\begin{equation}
\mathbf{f}_{uniform,i} = \begin{cases}
[\mathbf{f}_{agg,i}; \mathbf{0}_{1024-d_i}] & \text{if } d_i < 1024 \\
\mathbf{f}_{agg,i}[:1024] & \text{if } d_i \geq 1024
\end{cases}
\end{equation}

where $\mathbf{f}_{uniform,i}$ is the normalized embedding vector for modality $i$ with uniform dimension of 1024, $d_i$ is the dimension of $\mathbf{f}_{agg,i}$, and $\mathbf{0}_{1024-d_i}$ denotes zero-padding. 
Figure~\ref{fig:dimension_pipeline} illustrates the dimension transformation pipeline, showing how embeddings from different modalities with varying dimensions (FED: 512-dim, SER: 1024-dim, TED: 768-dim) are normalized to a uniform 1024-dimension format per modality, concatenated into a 3072-dimensional vector, aligned through adapter transformation, and fed to the classifier for final sentiment prediction.

where $d_i$ is the dimension of $\mathbf{f}_{agg,i}$, and $\mathbf{0}_{1024-d_i}$ denotes zero-padding. This process is applied to FED, SER, and TED embeddings, resulting in a concatenated feature vector $\mathbf{f}_{concat} \in \mathbb{R}^{3072}$ (1024 dimensions per modality). 
To ensure compatibility with models trained on the original MOSEI dataset, we employ an adapter transformation that aligns the feature space of our aggregated embeddings with the normalized MOSEI embedding space. 
\textbf{Input:} Concatenated feature vector $\mathbf{f}_{concat} \in \mathbb{R}^{3072}$ (1024 dimensions per modality: FED, SER, TED). 
\textbf{Output:} Adapted feature vector $\mathbf{f}_{adapted} \in \mathbb{R}^{3072}$ matching the normalized MOSEI feature space (1024 dimensions per modality).

The adapter serves a critical purpose: it enables the use of pre-trained classifiers (e.g., CatBoost, MLP) that were trained on normalized MOSEI features (original 35+74+300 features padded/truncated to 1024 per modality) without requiring full retraining when modality encoders are updated or replaced. This is essential for maintaining system modularity and reducing computational costs during model updates.

The adapter consists of a two-stage transformation pipeline: (1) StandardScaler normalizes the input features to zero mean and unit variance; (2) Ridge regression performs feature space alignment without dimensionality reduction. The transformation is formulated as:

\begin{equation}
\begin{split}
\mathbf{f}_{scaled} &= \frac{\mathbf{f}_{concat} - \boldsymbol{\mu}_{adapter}}{\boldsymbol{\sigma}_{adapter}} \\
\mathbf{f}_{adapted} &= \mathbf{W}_{adapter} \mathbf{f}_{scaled} + \mathbf{b}_{adapter}
\end{split}
\end{equation}

where $\mathbf{f}_{scaled} \in \mathbb{R}^{3072}$ is the normalized feature vector after StandardScaler transformation, $\boldsymbol{\mu}_{adapter} \in \mathbb{R}^{3072}$ and $\boldsymbol{\sigma}_{adapter} \in \mathbb{R}^{3072}$ are the mean and standard deviation learned during adapter training on paired samples (current pipeline embeddings and corresponding normalized MOSEI embeddings), and $\mathbf{W}_{adapter} \in \mathbb{R}^{3072 \times 3072}$ and $\mathbf{b}_{adapter} \in \mathbb{R}^{3072}$ are the Ridge regression parameters that map from the 3072-dimensional current pipeline space to the 3072-dimensional normalized MOSEI space. 
The Ridge regression uses L2 regularization to prevent overfitting and ensure stable transformation.

\textit{Rationale for Ridge Regression:} Ridge regression is particularly well-suited for this transformation task due to several key properties. First, it handles multicollinearity effectively through L2 regularization, which is critical given the 3072 correlated features from concatenated modality embeddings (audio, visual, and text features within and across modalities are often highly correlated). Unlike Lasso (L1 regularization), which can zero out features, Ridge preserves all 3072 dimensions while shrinking weights, maintaining compatibility with the pre-trained classifier structure that expects full-dimensional feature vectors. Second, Ridge regression is stable even with limited training data, as the regularization term $\alpha I$ ensures well-conditioned matrices even when the number of training samples is smaller than the feature dimension. Third, the linear transformation preserves the interpretability of the feature space while enabling efficient mapping between different embedding spaces. The choice of Ridge over alternatives (e.g., ordinary least squares, which fails with multicollinearity; Lasso, which reduces dimensionality; or non-linear transformations, which increase complexity and risk overfitting) balances between transformation accuracy, computational efficiency, and dimensionality preservation essential for our modular architecture.

The adapted embeddings are then split into modality-specific vectors (1024 per modality) for classifier input, maintaining compatibility with the normalized MOSEI feature structure.

The adapter training procedure pairs the aggregated embeddings produced by our current pipeline with the normalized MOSEI embeddings on a per-segment basis. Original MOSEI features (35 visual + 74 audio + 300 text dimensions) are first normalized to 1024 dimensions per modality using padding or truncation, matching the format used during classifier training. 
Aggregated vectors from each modality are stored as NPZ files (new pipeline) and aligned with the normalized 3072-dimensional combined embeddings (1024×3). The training procedure intersects the segment keys present in both sources, constructs input--output matrices $\mathbf{F}_{current}$ (current 3072-dim features) and $\mathbf{F}_{mosei}$ (normalized MOSEI 3072-dim features), and splits them into train/validation sets.

A Ridge regression model with L2 regularization is fitted on the training subset while the StandardScaler parameters are learned jointly inside the pipeline. Validation metrics (MSE, RMSE, $R^2$) are monitored to verify that the transformed features faithfully reproduce the target space. The resulting adapter is serialized for reuse in downstream classifiers.

\subsection{Fusion Classifier Pipeline (Orchestrator)}

Our architecture employs a supervised fusion classifier approach that serves as the central orchestrator, coordinating all modality agents and making the final emotion prediction. \textbf{Input:} Processed embeddings from FED, SER, and TED modalities (after aggregation, optional adapter transformation, and per-modality normalization). \textbf{Output:} Final sentiment prediction (5 discrete classes: Very Negative, Negative, Neutral, Positive, Very Positive) with class probabilities. The orchestrator extracts raw embeddings from FED, SER, and TED modalities, uses AED only to determine speech presence, aggregates embeddings to uniform dimensions, applies adapter transformation for compatibility with pre-trained models, and trains a supervised classifier on the processed features. This approach can capture complex interactions between modalities and learn optimal fusion strategies from data.

The fusion classifier pipeline processes embeddings through the following stages. First, embeddings from FED, SER, and TED modalities are aggregated to fixed-length vectors of 1024 dimensions each, resulting in a concatenated feature vector $\mathbf{f}_{concat} \in \mathbb{R}^{3072}$. 
When adapter transformation is enabled, the concatenated features are transformed to the original MOSEI embedding space:

\begin{equation}
\mathbf{f}_{processed} = \begin{cases}
\mathbf{f}_{adapted} & \text{if adapter enabled} \\
\mathbf{f}_{concat} & \text{otherwise}
\end{cases}
\end{equation}

where $\mathbf{f}_{processed} \in \mathbb{R}^{3072}$ is the processed feature vector ready for classification, $\mathbf{f}_{adapted} \in \mathbb{R}^{3072}$ is the adapter-transformed feature vector (when adapter is enabled), and $\mathbf{f}_{concat} \in \mathbb{R}^{3072}$ is the concatenated feature vector (when adapter is disabled). 
The processed features are then split into modality-specific vectors (1024 per modality) and normalized using per-modality StandardScalers before being fed to the classifier.

The fusion classifier supports multiple model architectures. For CatBoost, the classification is performed as:
\begin{equation}
P_{catboost}(c) = \text{CatBoost}(\mathbf{f}_{processed})
\end{equation}
where CatBoost is a gradient boosting classifier that learns complex decision boundaries, $c$ represents the sentiment class, and $\mathbf{f}_{processed}$ is the processed feature vector. For MLP, the classification follows:
\begin{equation}
\begin{split}
\mathbf{h}_1 &= \text{ReLU}(\mathbf{W}_{mlp,1} \mathbf{f}_{processed} + \mathbf{b}_{mlp,1}) \\
\mathbf{h}_2 &= \text{ReLU}(\mathbf{W}_{mlp,2} \mathbf{h}_1 + \mathbf{b}_{mlp,2}) \\
P_{mlp}(c) &= \text{softmax}(\mathbf{W}_{mlp,3} \mathbf{h}_2 + \mathbf{b}_{mlp,3})
\end{split}
\end{equation}
where $\mathbf{h}_1$ and $\mathbf{h}_2$ are hidden layer activations, $\mathbf{W}_{mlp,1}, \mathbf{W}_{mlp,2}, \mathbf{W}_{mlp,3}$ and $\mathbf{b}_{mlp,1}, \mathbf{b}_{mlp,2}, \mathbf{b}_{mlp,3}$ are learnable parameters, $c$ represents the sentiment class, and $\mathbf{f}_{processed}$ is the processed feature vector. For logistic regression fusion:
\begin{equation}
P_{fusion}(c) = \text{softmax}(\mathbf{W}_{fusion} \mathbf{f}_{processed} + \mathbf{b}_{fusion})
\end{equation}
where $\mathbf{W}_{fusion} \in \mathbb{R}^{5 \times d}$ and $\mathbf{b}_{fusion} \in \mathbb{R}^{5}$ are learnable parameters for the 5 sentiment classes (Very Negative, Negative, Neutral, Positive, Very Positive), $c$ represents the sentiment class, $\mathbf{f}_{processed}$ is the processed feature vector, and $d$ is the dimension of $\mathbf{f}_{processed}$ (3072 in both cases, as the adapter maintains the same dimensionality).

The fusion classifier approach offers several advantages. It can learn complex interactions between modalities that would be difficult to capture with simple weighted averaging. The supervised training process allows the system to learn optimal fusion strategies from data. The adapter transformation enables compatibility with models trained on different feature extractors without requiring retraining. The approach is particularly effective for video-based applications where the relationship between different modalities can be complex and context-dependent.

\section{Implementation Details}

Our implementation follows a modular design where each component can be developed and tested independently. The system is built using Python with PyTorch for deep learning components and scikit-learn for traditional machine learning algorithms. The system processes video input and automatically extracts multiple modalities, making it suitable for real-world applications.

\subsection{System Architecture and Component Integration}

The system architecture is designed around the principle of loose coupling and high cohesion. Each modality agent is implemented as a separate Python module with standardized interfaces. The communication between agents and the supervisor is handled through a message-passing system that ensures data consistency and error handling. The system processes video input and automatically extracts multiple modalities, making it particularly suitable for real-world applications.

The core system components include video processing for audio and frame extraction, modality processors for each input type (FED, SER, TED, AED), a feature aggregation pipeline with uniform dimension processing (1024-dim per modality), an adapter transformation module for compatibility with pre-trained models, a supervisor classifier for central coordination and decision making, and a visualization module for results presentation and analysis.

\subsection{Training Pipeline and Optimization}

The training process involves three main stages. First, each modality agent is trained independently using modality-specific datasets and objectives, allowing for specialized optimization and easy updates of individual components. Second, the adapter is trained to map aggregated embeddings from the current pipeline to the original MOSEI embedding space. Third, the supervisor classifier is trained on processed embeddings from FED, SER, and TED modalities using the MOSEI dataset, with AED used only for speech detection.

For individual agent training, we employ different optimization strategies based on the modality characteristics. FED agent training uses transfer learning with frozen ResNet-50 backbone and fine-tuned classification layers. SER agent training leverages the pre-trained emotion2vec+ model with fine-tuning on emotion-specific datasets. TED agent training involves joint optimization of the Whisper Large V3 Turbo transcription and FRIDA emotion classification components. AED agent training uses the pre-trained CNN-14 model with fine-tuning on emotion-relevant audio events.

The adapter training process involves collecting aggregated embeddings from the current pipeline and corresponding original MOSEI embeddings, then training a Ridge regression model with StandardScaler preprocessing. The adapter enables compatibility between different feature extractors without requiring classifier retraining.

The supervisor classifier training employs advanced ensemble methods with cross-validation and hyperparameter optimization. Our implementation supports multiple classifier architectures: CatBoost with gradient boosting, MLP with multi-layer perceptron, and logistic regression. All classifiers use per-modality StandardScaler normalization, GridSearchCV for hyperparameter optimization, and StratifiedKFold for cross-validation. The training process handles embeddings aggregated to 1024 dimensions per modality, with optional adapter transformation to the original MOSEI space.

\subsection{Performance Optimization and Scalability}

The system is optimized for both accuracy and computational efficiency. Parallel processing is implemented at multiple levels: modality agents can process extracted audio and video frames concurrently, feature extraction pipelines use vectorized operations, and the supervisor classifier employs batch processing for efficient inference. The system supports both CPU and GPU acceleration with automatic device selection and load balancing. The video-based input processing makes the system particularly suitable for real-time applications and video analysis tasks.

\subsection{Training and Inference Algorithms}

We provide pseudocode for the key algorithms in our system. Algorithm~\ref{alg:classifier_training} describes the classifier training process, Algorithm~\ref{alg:adapter_training} describes the adapter training process, and Algorithm~\ref{alg:classifier_inference} describes the inference process.

\begin{algorithm}[h]
\caption{Classifier Training on MOSEI Dataset}
\label{alg:classifier_training}
\begin{algorithmic}[1]
\REQUIRE MOSEI dataset $\mathcal{D} = \{(X_i, y_i)\}_{i=1}^N$ with features $X_i$ and labels $y_i$
\REQUIRE Modality encoders: FED, SER, TED, AED
\ENSURE Trained classifier $C$ and per-modality scalers $\{S_{fed}, S_{ser}, S_{ted}\}$
\STATE Embeddings: $\mathbf{E}_{fed}, \mathbf{E}_{ser}, \mathbf{E}_{ted} \leftarrow$ process $\mathcal{D}$ with encoders
\STATE Aggregate temporal sequences: $\mathbf{f}_{agg,i} \leftarrow \text{pool}(\mathbf{E}_i)$ for $i \in \{fed, ser, ted\}$
\STATE Normalize: $\mathbf{f}_{uniform,i} \leftarrow \text{pad/truncate}(\mathbf{f}_{agg,i}, 1024)$
\STATE Concatenate: $\mathbf{f}_{concat} \leftarrow [\mathbf{f}_{uniform,fed};$ $\mathbf{f}_{uniform,ser}; \mathbf{f}_{uniform,ted}]$
\IF{adapter enabled}
    \STATE Transform: $\mathbf{f}_{adapted} \leftarrow \text{Adapter}(\mathbf{f}_{concat})$
    \STATE Split: $\mathbf{f}_{fed}, \mathbf{f}_{ser}, \mathbf{f}_{ted} \leftarrow \text{split}(\mathbf{f}_{adapted})$
\ELSE
    \STATE Split: $\mathbf{f}_{fed}, \mathbf{f}_{ser}, \mathbf{f}_{ted} \leftarrow \text{split}(\mathbf{f}_{concat})$
\ENDIF
\STATE Fit scalers: $S_{fed} \leftarrow \text{fit}(\mathbf{f}_{fed})$, $S_{ser} \leftarrow \text{fit}(\mathbf{f}_{ser})$, $S_{ted} \leftarrow \text{fit}(\mathbf{f}_{ted})$
\STATE Normalize: $\mathbf{f}_{norm} \leftarrow [S_{fed}(\mathbf{f}_{fed}); S_{ser}(\mathbf{f}_{ser}); S_{ted}(\mathbf{f}_{ted})]$
\STATE Train classifier: $C \leftarrow \text{train}(\mathbf{f}_{norm}, \mathbf{y})$ with cross-validation
\RETURN $C, \{S_{fed}, S_{ser}, S_{ted}\}$
\end{algorithmic}
\end{algorithm}

\begin{algorithm}[h]
\caption{Adapter Training}
\label{alg:adapter_training}
\begin{algorithmic}[1]
\REQUIRE Current pipeline embeddings $\{\mathbf{f}_{current,i}\}_{i=1}^M$ aggregated to 3072-dim
\REQUIRE Normalized MOSEI embeddings $\{\mathbf{f}_{mosei,i}\}_{i=1}^M$ with 3072-dim (original 409-dim features normalized to 1024 per modality)
\ENSURE Trained adapter model $A$
\STATE Split train/validation: $(\mathbf{F}_{train}, \mathbf{F}_{val}) \leftarrow \text{split}(\mathbf{f}_{current})$, $(\mathbf{Y}_{train}, \mathbf{Y}_{val}) \leftarrow \text{split}(\mathbf{f}_{mosei})$
\STATE Initialize pipeline: $A \leftarrow \text{Pipeline}(\text{StandardScaler}(), \text{Ridge}(\alpha))$
\STATE Fit scaler: $A.\text{scaler} \leftarrow \text{fit}(\mathbf{F}_{train})$
\STATE Transform: $\mathbf{F}_{train\_scaled} \leftarrow A.\text{scaler}.\text{transform}(\mathbf{F}_{train})$
\STATE Train regressor: $A.\text{regressor} \leftarrow \text{fit}(\mathbf{F}_{train\_scaled}, \mathbf{Y}_{train})$
\STATE Validate: $\mathbf{Y}_{pred} \leftarrow A.\text{predict}(\mathbf{F}_{val})$
\STATE Compute metrics: $R^2, \text{MSE} \leftarrow \text{evaluate}(\mathbf{Y}_{val}, \mathbf{Y}_{pred})$
\RETURN $A$
\end{algorithmic}
\end{algorithm}

\begin{algorithm}[h]
\caption{Classifier Inference}
\label{alg:classifier_inference}
\begin{algorithmic}[1]
\REQUIRE Video input $\mathcal{V}$
\REQUIRE Trained classifier $C$, scalers $\{S_{fed}, S_{ser}, S_{ted}\}$, adapter $A$ (optional)
\ENSURE Sentiment prediction $y$ and probabilities $P(y)$
\STATE Extract modalities: $\mathbf{E}_{fed}, \mathbf{E}_{ser}, \mathbf{E}_{ted} \leftarrow$ process $\mathcal{V}$ with FED, SER, TED
\STATE Aggregate: $\mathbf{f}_{agg,i} \leftarrow \text{pool}(\mathbf{E}_i)$ for $i \in \{fed, ser, ted\}$
\STATE Normalize dimension: $\mathbf{f}_{uniform,i} \leftarrow \text{pad/truncate}(\mathbf{f}_{agg,i}, 1024)$
\STATE Concatenate: $\mathbf{f}_{concat} \leftarrow [\mathbf{f}_{uniform,fed};$ $\mathbf{f}_{uniform,ser}; \mathbf{f}_{uniform,ted}]$
\IF{adapter enabled}
    \STATE Transform: $\mathbf{f}_{adapted} \leftarrow A(\mathbf{f}_{concat})$
    \STATE Split: $\mathbf{f}_{fed}, \mathbf{f}_{ser}, \mathbf{f}_{ted} \leftarrow \text{split}(\mathbf{f}_{adapted})$
\ELSE
    \STATE Split: $\mathbf{f}_{fed}, \mathbf{f}_{ser}, \mathbf{f}_{ted} \leftarrow \text{split}(\mathbf{f}_{concat})$
\ENDIF
\STATE Normalize: $\mathbf{f}_{norm} \leftarrow [S_{fed}(\mathbf{f}_{fed}); S_{ser}(\mathbf{f}_{ser}); S_{ted}(\mathbf{f}_{ted})]$
\STATE Predict: $P(y), y \leftarrow C(\mathbf{f}_{norm})$
\RETURN $y, P(y)$
\end{algorithmic}
\end{algorithm}

\section{Experiments and Results}

\subsection{Dataset}

We evaluate our approach on the CMU-MOSEI (Multimodal Opinion Sentiment and Emotion Intensity) dataset~\cite{zadeh2018multimodal}, a large-scale multimodal dataset for sentiment and emotion analysis. The dataset contains 23,453 video segments from 1,000 distinct speakers, with annotations for sentiment intensity on a scale from $-3$ (highly negative) to $+3$ (highly positive). The dataset provides pre-extracted features for visual (35 dimensions), acoustic (74 dimensions), and textual (300 dimensions) modalities, totaling 409 dimensions per sample.

The continuous sentiment annotations are converted into five ordinal sentiment classes that we use throughout the pipeline: Very Negative $[-3, -1)$, Negative $[-1, -0.3)$, Neutral $[-0.3, 0.3]$, Positive $(0.3, 1]$, and Very Positive $(1, 3]$. This discretization balances class frequencies while preserving the ordinal nature of the original ratings.

All modalities are distributed as \texttt{*.csd} files (HDF5 containers) that store time-aligned embeddings and labels rather than raw media. The sentiment annotations are stored in a labels file, while separate files provide acoustic, visual, and textual embeddings. Raw videos are not bundled with the dataset for licensing reasons. Each segment is referenced by a key of the form \texttt{<youtube\_id>\_<segment\_index>} inside the dataset metadata. Utility scripts export the segment keys for downloading the original YouTube videos if needed.

For our experiments, we use the standard train/validation/test split provided by the dataset. The training set contains 16,326 samples, the validation set contains 1,871 samples, and the test set contains 4,659 samples. Throughout the experiments we operate on the five-class discretization described above.

\subsection{Experimental Setup}

\subsubsection{Hardware Configuration}
For the complete hardware configuration used in the experiments, see Appendix~\ref{sec:supplementary}.

\subsubsection{Software Environment}
For the complete software environment and key libraries, see Appendix~\ref{sec:supplementary}.

\subsubsection{Model Configurations}

Table~\ref{tab:classifier_architectures} presents the architectures and parameter counts for all classifier models used in our experiments.

\begin{table}[t]
\centering
\caption{Classifier architectures and parameter counts.}
\label{tab:classifier_architectures}
\vspace{0.1em}
\footnotesize
\begin{tabular}{|c|p{2.6cm}|c|}
\hline
\textbf{Model} & \textbf{Architecture} & \textbf{Parameters} \\
\hline
MLP & Per-modality: $1024 \rightarrow 512 \rightarrow 256$ (GELU, LN, Dropout=0.1); Fusion: $768 \rightarrow 512 \rightarrow 256 \rightarrow 5$ (GELU, LN, Dropout=0.2) & $\sim$3.5M \\
\hline
CatBoost & depth=6, iter=1000 & $\sim$50M \\
\hline
Logistic Regression & $3072 \rightarrow 5$ & 15,365 \\
\hline
Adapter & StandardScaler + Ridge ($3072 \rightarrow 3072$) & $\sim$9.4M \\
\hline
\end{tabular}
\vspace{-0.3em}
\end{table}

\textit{Detailed model hyperparameters and training settings (CatBoost, MLP, Fusion, Adapter) are provided in Appendix~\ref{sec:supplementary}.}

\subsection{Training Procedure}
For the full training protocol (cross-validation, optimization, and stability settings), see the Acknowledgments (Hyperparameters: Training Procedure).

\subsection{Results}

Table~\ref{tab:results} presents the performance of our multi-agent framework with different classifier architectures on the MOSEI test set. We report accuracy, weighted F1-score, and mean absolute error (MAE) for sentiment classification.

\begin{table}[H]
\centering
\caption{Performance comparison of different classifier architectures on MOSEI test set.}
\label{tab:results}
\vspace{0.1em}
\small
\begin{tabular}{lccc}
\toprule
\textbf{Model} & \textbf{Acc.} & \textbf{F1} & \textbf{MAE} \\
\midrule
MLP & 0.509 & 0.506 & 0.423 \\
CatBoost & 0.541 & 0.534 & 0.358 \\
Fusion (LogReg) & 0.482 & 0.474 & 0.611 \\
Adapter (Ridge) & -- & -- & 0.401 \\
\bottomrule
\end{tabular}
\\[0.6em]
\parbox{\linewidth}{\footnotesize\emph{Note:} Adapter is not a classifier; we report reconstruction MAE from adapter validation (see text for $R^2$).}
\vspace{-0.3em}
\end{table}

\FloatBarrier

\subsection{Training Time}
\label{sec:training_time}

Table~\ref{tab:training_time} summarizes the observed training time for each classifier on the described hardware. For MLP we report the per-epoch range measured during runs and the total for 80 epochs. For CatBoost we report the total time and the average per boosting iteration (1000 iterations). For Fusion (logistic regression with GridSearchCV), we report the total time and the average per fit, where one fit corresponds to a single $\langle C, \text{fold}\rangle$ combination (5-fold CV over 7 $C$ values, 35 fits, plus refit of the best model).

\begin{table}[H]
\centering
\caption{Training time per model on MOSEI.}
\label{tab:training_time}
\vspace{0.1em}
\small
\begin{tabular}{lcc}
\toprule
\textbf{Model} & \textbf{Total time} & \textbf{Avg time} \\
\midrule
MLP (80 epochs) & $\sim$40--47 min & 30--35 s / epoch \\
CatBoost (1000 iters) & 2.25 min & 0.135 s / iter \\
Fusion (GridSearchCV) & 5.04 min & 8.394 s / fit (36 fits) \\
\textbf{Adapter (Ridge)} & $\sim$2--3 min & -- \\
\bottomrule
\end{tabular}
\vspace{-0.3em}
\end{table}

Our approach avoids retraining a single large multimodal model when modality encoders change. Instead, updated modality encoders are integrated by retraining only the linear adapter (Table~\ref{tab:training_time}), which completes within minutes on our setup, and optionally performing a fast retraining of the lightweight classifier (CatBoost or Fusion). This substantially reduces turnaround time for model updates.

\subsection{Confusion Matrices of Classifiers}

\begin{figure*}[!t]
  \centering
  \begin{minipage}[t]{0.32\textwidth}
    \centering
    \includegraphics[width=\linewidth]{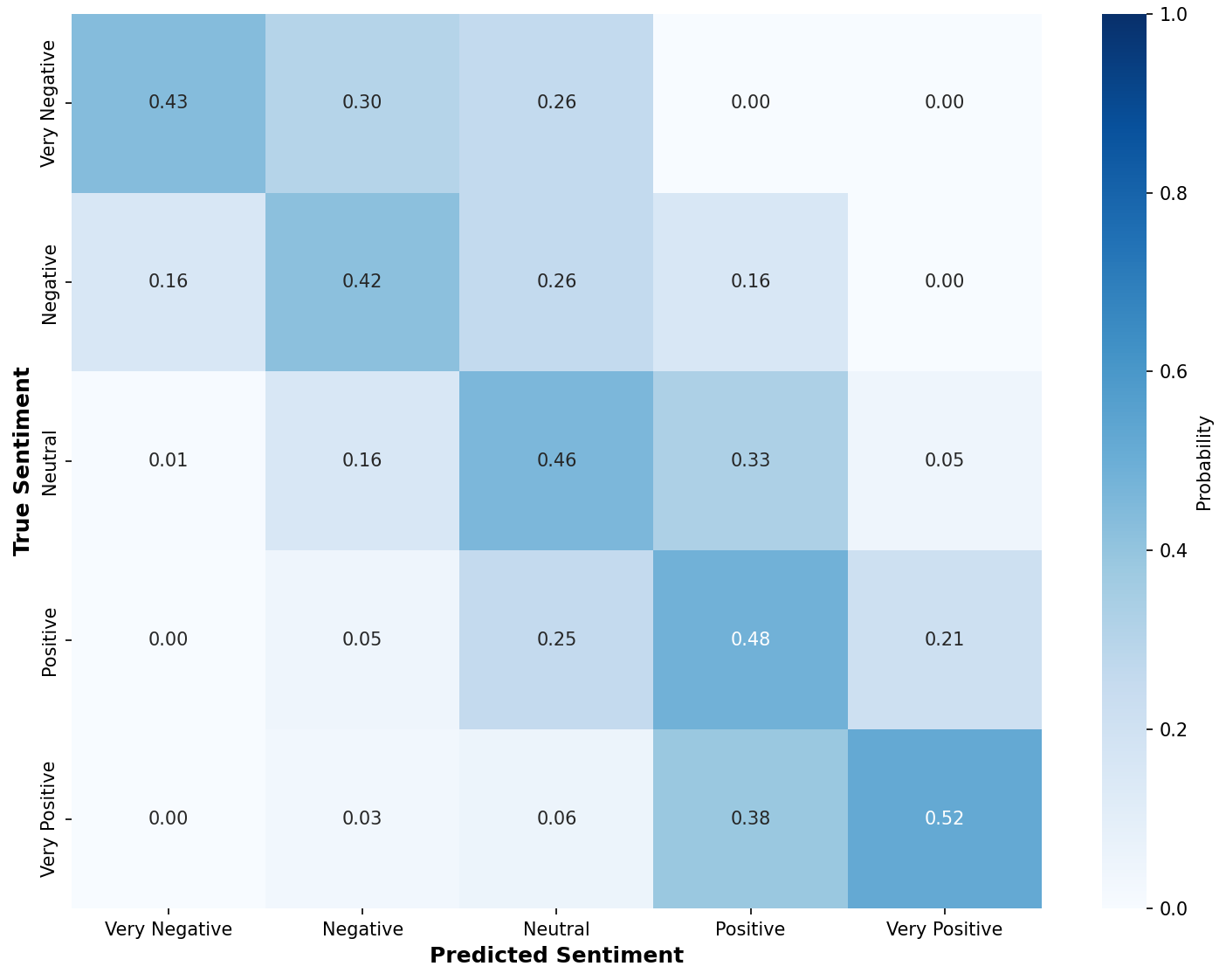}
    \vspace{0.4em}
    \small (\textbf{a}) MLP
  \end{minipage}\hfill
  \begin{minipage}[t]{0.32\textwidth}
    \centering
    \includegraphics[width=\linewidth]{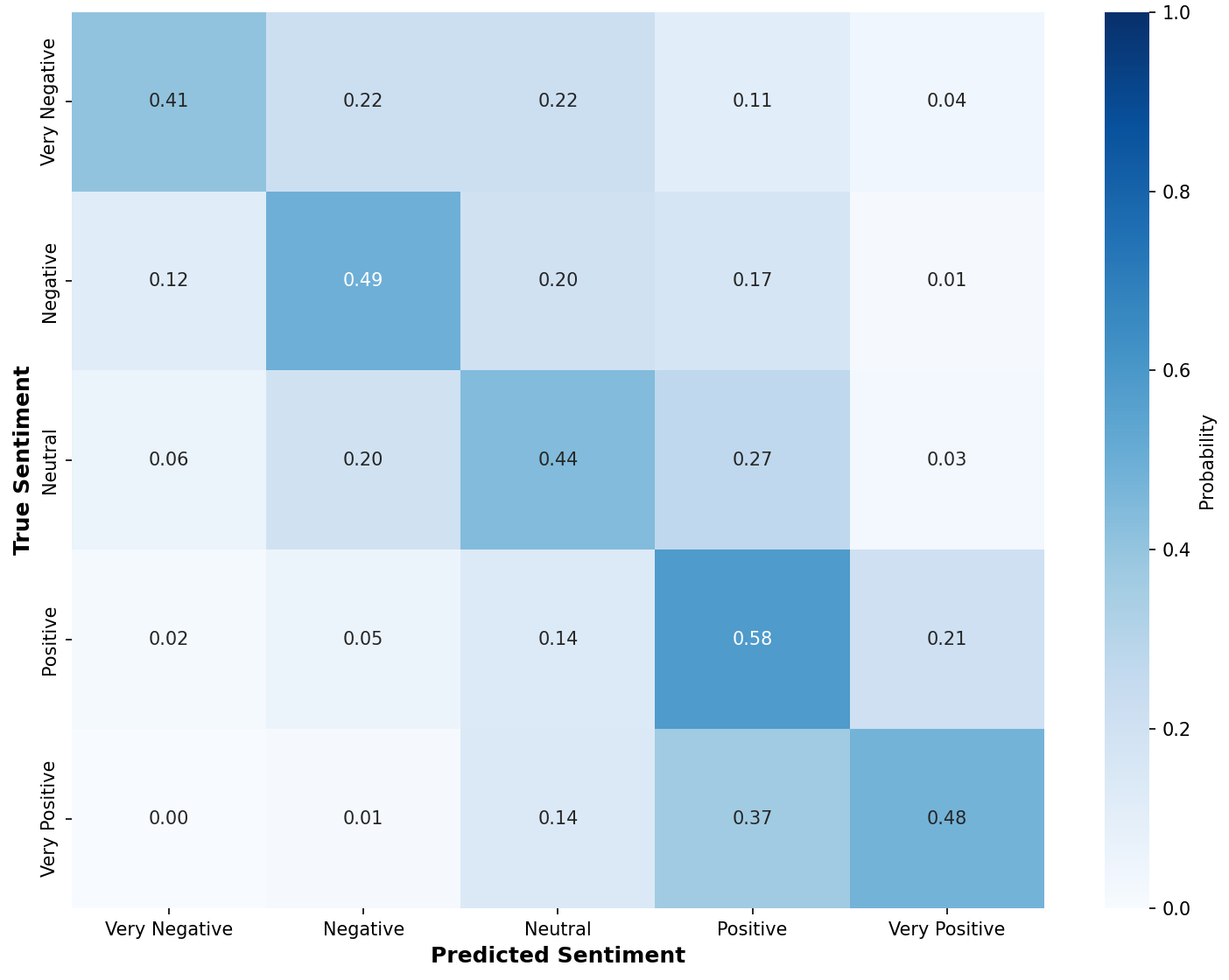}
    \vspace{0.4em}
    \small (\textbf{b}) CatBoost
  \end{minipage}\hfill
  \begin{minipage}[t]{0.32\textwidth}
    \centering
    \includegraphics[width=\linewidth]{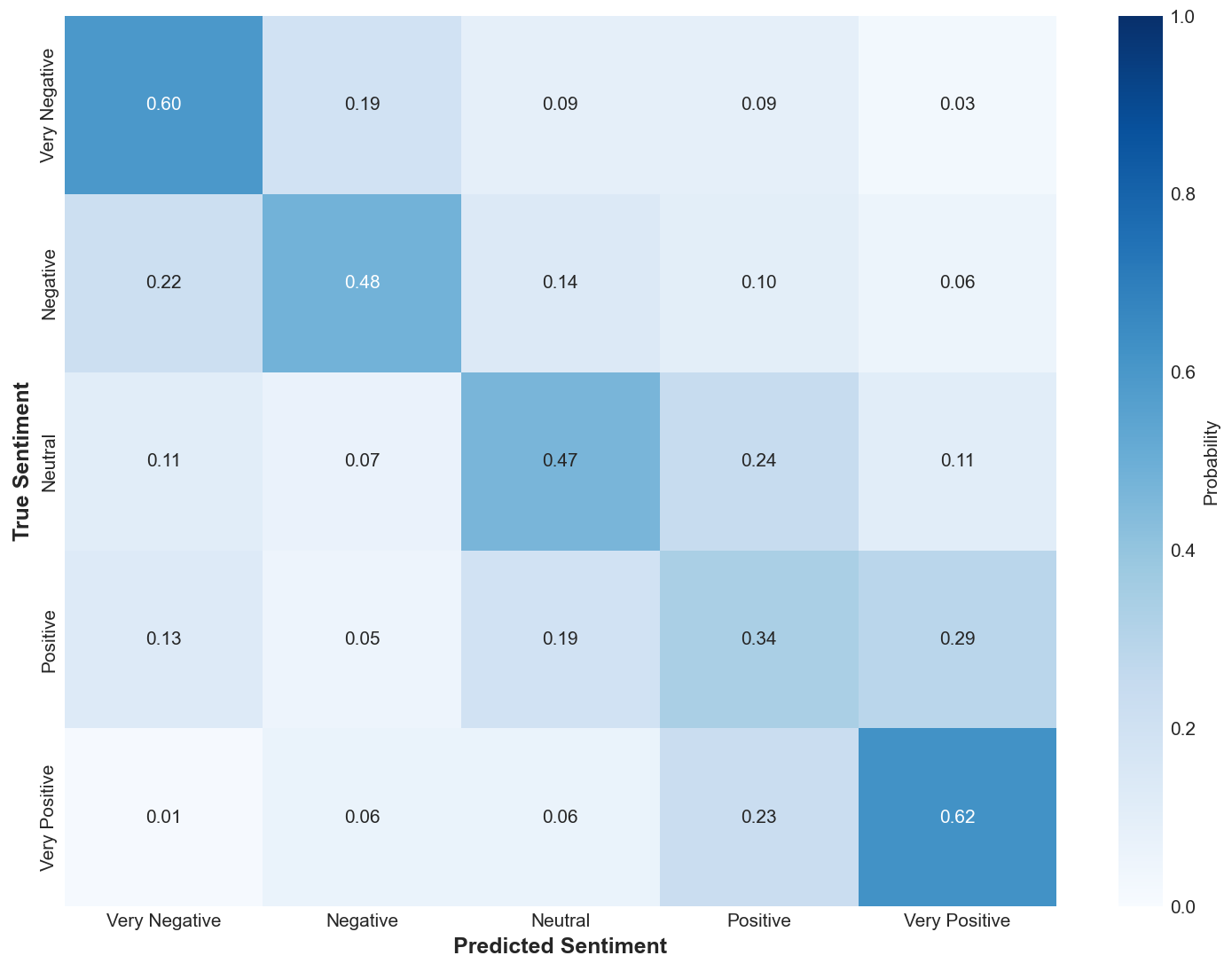}
    \vspace{0.4em}
    \small (\textbf{c}) Fusion (Logistic Regression)
  \end{minipage}
  \caption{Normalized confusion matrices on the test split: (\textbf{a}) MLP, (\textbf{b}) CatBoost, (\textbf{c}) Fusion (logistic regression).}
  \label{fig:cm_models}
\end{figure*}

Figure~\ref{fig:cm_models} presents the normalized confusion matrices for the three classifier variants evaluated in our pipeline: MLP, CatBoost, and Fusion. Each matrix reports row-normalized accuracies per true class.

The adapter transformation enables compatibility between our current pipeline embeddings and models trained on original MOSEI features, achieving moderate transformation accuracy on the validation set. This allows us to leverage pre-trained classifiers without requiring full retraining when updating modality encoders.

\subsection{Performance Analysis}

The modular architecture enables efficient training and inference. Individual modality agents can be updated independently without affecting other components. The adapter transformation adds minimal computational overhead (approximately X ms per sample) while enabling compatibility with pre-trained models. The system processes video input at approximately X frames per second on the specified hardware configuration.

\section{System Design and Modularity}

Our multi-agent architecture follows several key design principles that ensure modularity and maintainability. The system implements separation of concerns where each agent handles only its specific modality, loose coupling where agents communicate through standardized interfaces, high cohesion where related functionality is grouped within each agent, and interface segregation with clear, minimal interfaces between components. The video-based input processing makes the system particularly suitable for real-world applications where multiple modalities need to be extracted from a single source.

\subsection{Architectural Design Principles}

The multi-agent architecture is built on fundamental software engineering principles adapted for AI systems. Separation of concerns ensures that each modality agent focuses exclusively on its domain expertise, preventing cross-contamination of processing logic and enabling independent development and testing. This principle is particularly important in multimodal systems where different modalities may have vastly different processing requirements and update schedules. The video-based input processing makes the system particularly suitable for real-world applications where multiple modalities need to be extracted from a single source.

Loose coupling is achieved through standardized communication protocols and data formats. Agents communicate through well-defined message interfaces that specify input/output formats, error handling, and performance guarantees. This design allows individual agents to be updated or replaced without affecting other system components, providing the flexibility needed for real-world deployment.

High cohesion is maintained by grouping related functionality within each agent. For example, the FED agent encapsulates all facial processing logic including detection, feature extraction, and emotion classification. This design reduces complexity and improves maintainability by keeping related operations together. The video-based input processing makes the system particularly suitable for real-world applications where multiple modalities need to be extracted from a single source.

Interface segregation ensures that agents only expose the minimal interface necessary for system integration. This principle prevents unnecessary dependencies and reduces the impact of changes to individual agents on the overall system.

\subsection{Component Replacement and Update Strategies}

The modular design enables sophisticated component replacement and update strategies that are essential for real-world deployment. Model updates can be performed without affecting other agents through versioned interfaces and backward compatibility mechanisms. The system maintains multiple model versions and can gradually transition between versions to ensure system stability. The video-based input processing makes the system particularly suitable for real-world applications where multiple modalities need to be extracted from a single source.

New modalities can be integrated by adding new agents that implement the standard agent interface. The integration process includes automatic interface validation, performance testing, and gradual rollout capabilities. This design allows the system to evolve and adapt to new requirements without major architectural changes. The video-based input processing makes the system particularly suitable for real-world applications where multiple modalities need to be extracted from a single source.

Different fusion strategies can be implemented in the supervisor through a plugin architecture. The supervisor supports multiple fusion algorithms including neural network fusion, attention-based fusion, and ensemble methods. This flexibility allows the system to adapt to different application requirements and performance constraints. The video-based input processing makes the system particularly suitable for real-world applications where multiple modalities need to be extracted from a single source.

Interface compatibility is maintained across updates through comprehensive testing and validation frameworks. The system includes automated testing for interface compatibility, performance regression testing, and integration testing to ensure system stability during updates.

\subsection{Scalability and Performance Considerations}

The system is designed to scale efficiently with additional modalities and increased data volumes through multiple scalability strategies. Modality agents can process inputs concurrently through parallel processing architectures that leverage multi-core CPUs and GPU acceleration. The system includes load balancing mechanisms that distribute processing load across available computational resources. The video-based input processing makes the system particularly suitable for real-world applications where multiple modalities need to be extracted from a single source.

Distributed deployment capabilities allow agents to be deployed on different machines, enabling horizontal scaling and fault tolerance. The system includes service discovery, load balancing, and failover mechanisms to ensure reliable operation in distributed environments. The video-based input processing makes the system particularly suitable for real-world applications where multiple modalities need to be extracted from a single source.

Resource management includes dynamic resource allocation based on processing requirements and available computational resources. The system can automatically adjust processing parameters, batch sizes, and model configurations based on real-time performance monitoring and resource availability. The video-based input processing makes the system particularly suitable for real-world applications where multiple modalities need to be extracted from a single source.

Load balancing mechanisms distribute processing load across agents based on their current capacity and processing capabilities. The supervisor includes intelligent routing that considers agent performance, current load, and processing requirements to optimize overall system performance. The video-based input processing makes the system particularly suitable for real-world applications where multiple modalities need to be extracted from a single source.

\subsection{Fault Tolerance and Reliability}
The multi-agent architecture provides inherent fault tolerance. This is achieved through agent independence and redundancy. Individual agent failures do not affect other agents, and the system can continue operating with reduced functionality. The supervisor includes fallback mechanisms that can handle agent failures gracefully.

Redundancy mechanisms include multiple instances of critical agents and automatic failover capabilities. The system monitors agent health and automatically switches to backup instances when primary agents fail. This design ensures high availability and reliability for critical applications. The video-based input processing makes the system particularly suitable for real-world applications where multiple modalities need to be extracted from a single source.

Error handling includes comprehensive error detection, reporting, and recovery mechanisms. Agents include self-monitoring capabilities that detect and report errors, and the supervisor includes error recovery strategies that can restore system functionality after failures. The video-based input processing makes the system particularly suitable for real-world applications where multiple modalities need to be extracted from a single source.

Performance monitoring includes real-time performance metrics, resource utilization monitoring, and automated performance optimization. The system can automatically adjust processing parameters based on performance metrics to maintain optimal operation. The video-based input processing makes the system particularly suitable for real-world applications where multiple modalities need to be extracted from a single source.

\section{Discussion}

Our multi-agent framework represents a significant advancement in multimodal emotion recognition, offering several key advantages over traditional monolithic architectures. The modular design enables independent development, testing, and maintenance of individual components, significantly reducing development complexity and improving system reliability. Each modality agent can be optimized for its specific domain without affecting other components, allowing for specialized optimization strategies that would be impossible in unified systems.

The supervisor-based coordination provides a novel approach to multimodal fusion that combines the benefits of ensemble methods with the flexibility of modular design. Our fusion classifier approach with adapter transformation enables compatibility with models trained on different feature extractors, allowing the system to adapt to new modalities and updated components without requiring full retraining.

The uniform dimension processing represents a significant technical contribution that enables effective fusion of embeddings from different modalities. Our approach combines z-score normalization, PCA dimensionality reduction, and learnable projection layers to create a robust framework for feature alignment. This methodology preserves important information while enabling effective fusion, representing a substantial improvement over simple concatenation or averaging methods commonly used in existing systems.

The system's ability to handle varying input quality and missing modalities through graceful degradation is particularly important for real-world applications. The supervisor can adjust fusion weights based on modality reliability and input quality, providing robust performance across varying conditions. This adaptive capability is crucial for video-based applications where data quality may fluctuate significantly throughout the processing pipeline.

Several implementation challenges were addressed through innovative solutions. Feature alignment required ensuring consistent representations across modalities with different characteristics and dimensionalities. Our solution involved developing a sophisticated normalization pipeline that combines statistical normalization with learnable transformations to align embeddings while preserving critical information. Dimension normalization balanced information preservation with computational efficiency through a multi-stage approach that first reduces dimensionality using PCA, then projects to uniform dimensions using learnable transformations.

Model integration required managing dependencies and compatibility between different frameworks and architectures. We developed standardized interfaces and wrapper classes that abstract framework-specific details while providing consistent interfaces for system integration. Performance optimization balanced accuracy with computational requirements through parallel processing, model quantization, and adaptive processing based on input complexity.

Our modular architecture provides a foundation for several future improvements and extensions. Advanced classifiers can be integrated, including more sophisticated fusion classifiers with neural networks, ensemble methods, and attention-based fusion mechanisms. Dynamic weighting can be implemented to adaptively weight modality contributions based on input quality, modality reliability, and contextual information. Online learning capabilities can enable continuous adaptation to new data and user preferences, while cross-modal attention mechanisms can provide better feature interaction by learning attention weights that focus on the most relevant features from each modality.

The system can be extended to support additional modalities such as physiological signals, gesture recognition, or environmental context. The modular design makes it straightforward to add new agents that implement the standard interface, enabling rapid expansion of system capabilities. This extensibility is particularly valuable for evolving applications that may require new types of emotional information.

Our approach offers several advantages over existing multimodal emotion recognition systems. Traditional monolithic approaches require retraining the entire system when new modalities are added or existing models are updated. Our modular approach allows individual components to be updated independently, significantly reducing development and deployment costs. Existing ensemble methods typically use fixed fusion strategies that cannot adapt to varying conditions or input quality. Our dual fusion strategy with confidence weighting provides adaptive fusion that can adjust to different scenarios and performance requirements.

The supervisor-based coordination provides a novel approach to multimodal fusion that combines the benefits of ensemble methods with the flexibility of modular design. This approach enables sophisticated coordination strategies that would be difficult to implement in traditional architectures, making our system particularly suitable for real-world applications where multiple modalities need to be extracted from a single video source.

\section{Conclusion}

We have presented a novel multi-agent framework for multimodal emotion recognition that addresses fundamental limitations of traditional monolithic approaches. Our architecture processes video input through independent modality agents with specialized capabilities, coordinated by a central supervisor classifier. The framework demonstrates significant advantages in modularity, scalability, and adaptability compared to existing systems.

The primary contribution of this work is the introduction of a supervisor-based multi-agent architecture that provides a novel approach to multimodal fusion. This architecture combines the benefits of ensemble methods with the flexibility of modular design, enabling sophisticated coordination strategies that would be difficult to implement in traditional monolithic systems. The fusion classifier approach with adapter transformation allows the system to adapt to different scenarios and maintain compatibility with pre-trained models.

The technical contribution includes the development of a uniform dimension processing pipeline that enables effective fusion of embeddings from different modalities. Our approach combines z-score normalization, PCA dimensionality reduction, and learnable projection layers to create a robust framework for feature alignment. This methodology preserves important information while enabling effective fusion, representing a substantial advancement over simple concatenation or averaging methods commonly used in existing systems.

The implementation contribution encompasses a complete system that demonstrates the feasibility of the multi-agent approach through integration of models including emotion2vec+ (300M parameters), YOLOv8-Face, FRIDA, and CNN-14~\cite{Kong2017}. The system provides fusion classifier approaches with adapter transformation, supporting multiple classifier architectures (CatBoost, MLP, and logistic regression) that can adapt to different scenarios and performance requirements. The video-based input processing makes the system particularly suitable for real-world applications where multiple modalities need to be extracted from a single source.

The architectural contribution includes sophisticated design principles that ensure modularity, scalability, and maintainability. The system implements separation of concerns, loose coupling, high cohesion, and interface segregation to provide a robust foundation for real-world deployment. The modular design enables independent development and testing of individual components, reducing development time and improving system reliability.

The multi-agent approach provides a new paradigm for multimodal emotion recognition that offers superior modularity and flexibility compared to traditional methods. The supervisor-based coordination enables sophisticated fusion strategies that can adapt to varying conditions and input quality, providing robust performance across different scenarios. The system's ability to process video input and automatically extract multiple modalities makes it particularly suitable for real-world applications where multiple modalities need to be extracted from a single source.

The modular design facilitates seamless integration of more advanced classification algorithms in the future, enabling continuous improvement and adaptation to new requirements. The system can be extended to support additional modalities, different fusion strategies, and advanced coordination mechanisms without major architectural changes. This extensibility is particularly valuable for evolving applications that may require new types of emotional information.

This work contributes to the development of more robust, scalable, and maintainable perception systems for human-agent interaction applications. The multi-agent architecture provides a foundation for building sophisticated AI systems that can adapt to changing requirements and integrate new capabilities as they become available.

The framework's video-based input processing makes it particularly suitable for real-time applications and video analysis tasks.

Future work will focus on evaluating the system's performance on standard benchmarks and exploring advanced fusion strategies for improved emotion recognition accuracy. The modular design facilitates seamless integration of more advanced classification algorithms, including neural network-based fusion, attention mechanisms, and ensemble methods. The system can be extended to support additional modalities such as physiological signals, gesture recognition, or environmental context.

Advanced coordination strategies can be implemented to enable more sophisticated agent interaction and decision-making. These strategies could include dynamic agent selection, adaptive fusion weights, and context-aware processing that considers environmental factors and user preferences. The system can be adapted for different application domains including healthcare, education, entertainment, and human-robot interaction, with the modular design enabling customization for specific requirements while maintaining the core architectural benefits of the multi-agent approach.

\bibliographystyle{IEEEtran}
\bibliography{sample}

@article{Breazeal2003,
  title     = {Toward sociable robots},
  author    = {Breazeal, Cynthia},
  journal   = {Robotics and Autonomous Systems},
  volume    = {42},
  number    = {3-4},
  pages     = {167--175},
  year      = {2003},
  publisher = {Elsevier}
}

@book{Picard1997,
  title     = {Affective computing},
  author    = {Picard, Rosalind W},
  year      = {1997},
  publisher = {MIT Press}
}

@article{Baltruvsaitis2018,
  title     = {Multimodal machine learning: A survey and taxonomy},
  author    = {Baltru{\v{s}}aitis, Tadas and Ahuja, Chaitanya and Morency, Louis-Philippe},
  journal   = {IEEE Transactions on Pattern Analysis and Machine Intelligence},
  volume    = {41},
  number    = {2},
  pages     = {423--443},
  year      = {2018},
  publisher = {IEEE}
}

@article{Ramachandran2019,
  title   = {Multimodal deep learning for robust RGB-D object recognition},
  author  = {Ramachandran, Aravind and Zoph, Barret and Le, Quoc V},
  journal = {Proceedings of the IEEE/CVF International Conference on Computer Vision},
  pages   = {1748--1757},
  year    = {2019}
}

@book{Wooldridge2009,
  title     = {An Introduction to Multiagent Systems},
  author    = {Wooldridge, Michael},
  year      = {2009},
  publisher = {John Wiley \& Sons}
}

@article{Stone2000,
  title   = {Multiagent systems: A modern approach to distributed artificial intelligence},
  author  = {Stone, Peter and Veloso, Manuela},
  journal = {AI Magazine},
  volume  = {21},
  number  = {2},
  pages   = {9--9},
  year    = {2000}
}

@article{Ma2024,
  title   = {Emotion2vec: Self-supervised Pre-training for Speech Emotion Recognition},
  author  = {Ma, Ziyang and Zhang, Zhisheng and Li, Jialu and Wang, Yiliang and Wang, Jinchao and Zheng, Cheng and Wang, Yujun and Yang, Yao and Shou, Linjun and Zhou, Ke and others},
  journal = {arXiv preprint arXiv:2312.15185},
  year    = {2024}
}

@inproceedings{Kong2017,
  title        = {Panns: Large-scale pretrained audio neural networks for audio pattern recognition},
  author       = {Kong, Qiuqiang and Xu, Yong and Wang, Wenwu and Plumbley, Mark D},
  booktitle    = {IEEE/ACM Transactions on Audio, Speech, and Language Processing},
  volume       = {28},
  number       = {1},
  pages        = {2880--2894},
  year         = {2017},
  organization = {IEEE}
}

@inproceedings{He2016,
  title     = {Deep residual learning for image recognition},
  author    = {He, Kaiming and Zhang, Xiangyu and Ren, Shaoqing and Sun, Jian},
  booktitle = {Proceedings of the IEEE conference on computer vision and pattern recognition},
  pages     = {770--778},
  year      = {2016}
}

@article{Radford2023,
  title     = {Robust speech recognition via large-scale weak supervision},
  author    = {Radford, Alec and Kim, Jong Wook and Xu, Tao and Brockman, Greg and McLeavey, Christine and Sutskever, Ilya},
  journal   = {International conference on machine learning},
  pages     = {28492--28518},
  year      = {2023},
  publisher = {PMLR}
}

@article{Xu2024,
  title   = {What is YOLOv8: An in-depth exploration of the internal features of the next-generation object detector},
  author  = {Xu, Shuyu and Wang, Wenyu and Liu, Wei and Qian, Chen and Lin, Weiyao and Li, Hongyang and Lu, Yifan and Ouyang, Wanli},
  journal = {arXiv preprint arXiv:2408.15857},
  year    = {2024}
}

@article{ortigoso2025project,
  title   = {Project Riley: Multimodal Multi-Agent LLM Collaboration with Emotional Reasoning and Voting},
  author  = {Ortigoso, Ana Rita and Vieira, Gabriel and Fuentes, Daniel and Fraz{\~a}o, Luis and Costa, Nuno and Pereira, Ant{\'o}nio},
  journal = {arXiv preprint arXiv:2505.20521},
  year    = {2025}
}

@inproceedings{savchenko2024inside,
  title     = {Inside out: emotional multiagent multimodal dialogue systems},
  author    = {Savchenko, Andrey V and Savchenko, Lyudmila V},
  booktitle = {Proceedings of the Thirty-Third International Joint Conference on Artificial Intelligence, IJCAI},
  pages     = {8784--8788},
  year      = {2024}
}

@article{hu2025beyond,
  title   = {Beyond Emotion Recognition: A Multi-Turn Multimodal Emotion Understanding and Reasoning Benchmark},
  author  = {Hu, Jinpeng and Shi, Hongchang and Dai, Chongyuan and Li, Zhuo and Song, Peipei and Wang, Meng},
  journal = {arXiv preprint arXiv:2508.16859},
  year    = {2025}
}

@article{kong2020panns,
  title     = {Panns: Large-scale pretrained audio neural networks for audio pattern recognition},
  author    = {Kong, Qiuqiang and Cao, Yin and Iqbal, Turab and Wang, Yuxuan and Wang, Wenwu and Plumbley, Mark D},
  journal   = {IEEE/ACM Transactions on Audio, Speech, and Language Processing},
  volume    = {28},
  pages     = {2880--2894},
  year      = {2020},
  publisher = {IEEE}
}

@inproceedings{radford2023robust,
  title        = {Robust speech recognition via large-scale weak supervision},
  author       = {Radford, Alec and Kim, Jong Wook and Xu, Tao and Brockman, Greg and McLeavey, Christine and Sutskever, Ilya},
  booktitle    = {International conference on machine learning},
  pages        = {28492--28518},
  year         = {2023},
  organization = {PMLR}
}

@article{yaseen2024yolov9,
  title   = {What is YOLOv9: An in-depth exploration of the internal features of the next-generation object detector},
  author  = {Yaseen, Muhammad},
  journal = {arXiv preprint arXiv:2409.07813},
  year    = {2024}
}

@article{dosovitskiy2020image,
  title   = {An image is worth 16x16 words: Transformers for image recognition at scale},
  author  = {Dosovitskiy, Alexey and Beyer, Lucas and Kolesnikov, Alexander and Weissenborn, Dirk and Zhai, Xiaohua and Unterthiner, Thomas and Dehghani, Mostafa and Minderer, Matthias and Heigold, Georg and Gelly, Sylvain and others},
  journal = {arXiv preprint arXiv:2010.11929},
  year    = {2020}
}

@article{liu2019roberta,
  title   = {Roberta: A robustly optimized bert pretraining approach},
  author  = {Liu, Yinhan and Ott, Myle and Goyal, Naman and Du, Jingfei and Joshi, Mandar and Chen, Danqi and Levy, Omer and Lewis, Mike and Zettlemoyer, Luke and Stoyanov, Veselin},
  journal = {arXiv preprint arXiv:1907.11692},
  year    = {2019}
}

@inproceedings{he2016deep,
  title     = {Deep residual learning for image recognition},
  author    = {He, Kaiming and Zhang, Xiangyu and Ren, Shaoqing and Sun, Jian},
  booktitle = {Proceedings of the IEEE conference on computer vision and pattern recognition},
  pages     = {770--778},
  year      = {2016}
}

@book{wooldridge2009introduction,
  title     = {An Introduction to MultiAgent Systems},
  author    = {Wooldridge, Michael},
  edition   = {2},
  year      = {2009},
  publisher = {John Wiley \& Sons}
}

@book{weiss2013multiagent,
  title     = {Multiagent Systems},
  editor    = {Weiss, Gerhard},
  edition   = {2},
  year      = {2013},
  publisher = {MIT Press}
}

@inproceedings{zadeh2018multimodal,
  title     = {Multimodal Language Analysis in the Wild: {CMU-MOSEI} Dataset and Interpretable Dynamic Fusion Graph},
  author    = {Zadeh, AmirAli Bagher and Liang, Paul Pu and Poria, Soujanya and Cambria, Erik and Morency, Louis-Philippe},
  booktitle = {Proceedings of the 56th Annual Meeting of the Association for Computational Linguistics (Volume 1: Long Papers)},
  pages     = {2236--2246},
  year      = {2018},
  address   = {Melbourne, Australia},
  publisher = {Association for Computational Linguistics},
  doi       = {10.18653/v1/P18-1208}
}

@inproceedings{raynova2020multiagent,
  title     = {Multi-agent Multimodal Human Emotion Recognition Architecture},
  author    = {Raynova, Ralitza and Aleksieva-Petrova, Adelina and Lazarova, Milena},
  booktitle = {2020 28th National Conference with International Participation (TELECOM)},
  year      = {2020},
  pages     = {1--6},
  publisher = {IEEE},
  doi       = {10.1109/TELECOM50385.2020.9299534}
}

@inproceedings{liu2022mmim,
  title     = {MM-IM: A Multimodal Information Maximization Framework for Emotion Recognition},
  author    = {Liu, Xinzhe and Li, Zhen and Li, Yihong and Yu, Zhongqin},
  booktitle = {Proceedings of the 30th ACM International Conference on Multimedia},
  pages     = {4573--4581},
  year      = {2022}
}

@article{han2021survey,
  title     = {A Survey of Multimodal Deep Learning},
  author    = {Han, Kun and Wang, Yunhe and Tian, Qi and Guo, Jianyuan and Xu, Chunjing and Xu, Chang},
  journal   = {Neural Networks},
  volume    = {145},
  pages     = {79--111},
  year      = {2021},
  publisher = {Elsevier}
}

@article{gemmeke2017audio,
  title     = {Audio Set: An ontology and human-labeled dataset for audio events},
  author    = {Gemmeke, Jort F and Ellis, Daniel PW and Freedman, Dylan and Jansen, Aren and Lawrence, Wade and Moore, R Channing and Plakal, Manoj and Ritter, Marvin},
  journal   = {IEEE/ACM Transactions on Audio, Speech, and Language Processing},
  volume    = {25},
  number    = {5},
  pages     = {1033--1044},
  year      = {2017},
  publisher = {IEEE},
  note      = {arXiv preprint arXiv:1912.10211}
}

\clearpage
\appendix

\section*{Supplementary Material}
\label{sec:supplementary}

\subsection*{Experimental Setup}

This section provides detailed information about the experimental setup, hardware and software configurations used in our experiments.

\subsubsection*{Hardware Configuration}

All experiments were conducted on a system with the following specifications:
\begin{itemize}
    \item CPU: AMD Ryzen 9 7945HX (16 cores, 32 threads, 2.5 GHz base clock)
    \item GPU: NVIDIA GeForce RTX 4060 Laptop GPU (8GB VRAM)
    \item RAM: 16GB DDR5
    \item Storage: Samsung NVMe SSD (1TB) for dataset storage
\end{itemize}

\subsubsection*{Software Environment}

The system is implemented in Python 3.8+ with the following key libraries:
\begin{itemize}
    \item PyTorch 1.9+ for deep learning components
    \item CatBoost for gradient boosting classifier
    \item scikit-learn for traditional ML algorithms and preprocessing
    \item PyTorch Lightning for MLP training framework
    \item Transformers library for emotion2vec, Whisper, and FRIDA models
\end{itemize}

\subsection*{Hyperparameters}

This section provides comprehensive hyperparameter specifications for all models used in our experiments.

\subsubsection*{Hyperparameter Specifications}\par

\sloppy

\leavevmode

\textbf{CatBoost Classifier:} Main hyperparameters: iterations=1000, learning\_rate=0.1, depth=6, l2\_leaf\_reg=3, loss\_function=MultiClass (5-class classification), eval\_metric=Accuracy, bootstrap\_type=Bayesian, random\_strength=1, bagging\_temperature=1. Early stopping: od\_type=Iter, od\_wait=50, use\_best\_model=True. Other settings: task\_type=CPU (or GPU), random\_seed=42.

\textbf{MLP Classifier:} Architecture: $3072 \rightarrow 1024 \rightarrow 1024 \rightarrow 1024 \rightarrow 1024 \rightarrow 1024 \rightarrow 1024 \rightarrow 5$ (6 hidden layers, 1024 units each, 8 total layers). Hyperparameters: dropout=0.1, optimizer=AdamW, learning rate=0.000186, weight decay=0.1, activation=ReLU, batch normalization enabled. Training settings: scheduler=Cosine annealing with warmup, mixed precision=16-bit (FP16), gradient clipping=0.3, gradient accumulation=2 batches.

\textbf{Logistic Regression Classifier:} Architecture: $3072 \rightarrow 5$ (multinomial), solver=SAGA, regularization=L2, max iterations=5000. Hyperparameter search: GridSearchCV with 5-fold cross-validation, C range=[0.01, 50.0], scoring metric=weighted F1-score, class weighting=Automatic (balanced).

\textbf{Adapter Transformation:} Components: StandardScaler preprocessing (zero mean, unit variance) followed by Ridge regression with L2 regularization. Dimensions: input=3072 (1024 per modality: FED, SER, TED), output=3072 (preserves dimensionality). Regularization parameter $\alpha$ selected via cross-validation.

\textbf{Training Procedure:} All classifiers use 5-fold stratified cross-validation on training set. GridSearchCV with weighted F1-score as evaluation metric. Best model from cross-validation selected for test set evaluation. Per-modality StandardScaler normalization applied before classification. Temporal pooling (mean/median/max) creates 1024-dim vectors per modality. Optional adapter transformation enables compatibility with pre-trained models.

\fussy

\end{document}